%% file: iclr2026_conference.tex
\documentclass{article} 
\usepackage{iclr2026_conference,times}

\input{math_commands.tex}

\usepackage{float}
\usepackage{amsmath,amssymb,amsfonts}
\usepackage{algorithmic}
\usepackage{algorithm}
\usepackage{graphicx}
\usepackage{textcomp}
\usepackage{xcolor}
\usepackage{booktabs} 
\usepackage{hyperref} 
\usepackage{tikz}
\usepackage{natbib}
\usetikzlibrary{positioning}

\def\BibTeX{{\rm B\kern-.05em{\sc i\kern-.025em b}\kern-.08em
    T\kern-.1667em\lower.7ex\hbox{E}\kern-.125emX}}

\iclrfinalcopy

\begin{document}

\title{LASER: Low-Rank Activation SVD for Efficient Recursion}

\author{Ege Çakar, Ketan Raghu \& Lia Zheng \\
\textit{Harvard University} \\
Cambridge, MA, USA \\
\textit{\{ecakar, karaghu, liazheng\}@college.harvard.edu}
}

\maketitle

\begin{abstract}
Recursive architectures such as Tiny Recursive Models (TRMs)
perform implicit reasoning through iterative latent computation,
yet the geometric structure of these reasoning trajectories
remains poorly understood. We investigate the activation manifold
of TRMs during recursive unrolling and find that activations
occupy an effectively linear, low-dimensional subspace whose
principal directions can be tracked dynamically with cheap power
iterations. This suggests that weight-sharing concentrates
iterative computation along a small number of dominant
eigendirections, and we find that this concentration varies
sharply across computational sites. We exploit this structure
through LASER (Low-Rank Activation SVD for Efficient Recursion),
a dynamic compression framework that maintains an evolving
low-rank basis via matrix-free subspace tracking with a
fidelity-triggered reset mechanism, achieving ${\sim}60\%$
activation memory savings with no statistically significant
accuracy degradation. Our analysis raises questions about how
recursive architectures allocate representational capacity during
implicit reasoning, and whether this concentration can be
exploited to improve the efficiency and stability of latent
computation.
\end{abstract}

\begin{figure}[h!]
\centering
\begin{tikzpicture}[scale=0.9, every node/.style={font=\small}]

  \node[font=\bfseries] at (0, 2.2) {Standard TRM};
  \node[draw, fill=red!20, minimum width=2.8cm, minimum height=0.55cm] at (0, 1.4) {$X_1 \in \mathbb{R}^{B \times D}$};
  \node[draw, fill=red!20, minimum width=2.8cm, minimum height=0.55cm] at (0, 0.7) {$X_2 \in \mathbb{R}^{B \times D}$};
  \node at (0, 0.1) {$\vdots$};
  \node[draw, fill=red!20, minimum width=2.8cm, minimum height=0.55cm] at (0, -0.5) {$X_n \in \mathbb{R}^{B \times D}$};
  \node[font=\scriptsize, text=red!70!black] at (0, -1.2) {Memory: $O(nBD)$};

  \draw[->, thick] (1.9, 0.5) -- (3.3, 0.5);
  \node[font=\scriptsize] at (2.6, 0.8) {LASER};

  \node[font=\bfseries] at (5.8, 2.2) {Compressed TRM};
  \node[draw, fill=green!20, minimum width=2.2cm, minimum height=0.55cm] at (4.7, 1.4) {$Z_1 \in \mathbb{R}^{B \times k}$};
  \node[draw, fill=green!20, minimum width=2.2cm, minimum height=0.55cm] at (4.7, 0.7) {$Z_2 \in \mathbb{R}^{B \times k}$};
  \node at (4.7, 0.1) {$\vdots$};
  \node[draw, fill=green!20, minimum width=2.2cm, minimum height=0.55cm] at (4.7, -0.5) {$Z_n \in \mathbb{R}^{B \times k}$};

  \node[draw, fill=blue!20, minimum width=1.6cm, minimum height=2.0cm] at (7.1, 0.45) {$Q \in \mathbb{R}^{D \times k}$};

  \node[font=\scriptsize] at (5.9, -1.2) {$k$ = retained rank, $k \ll D$};
  \node[font=\scriptsize, text=green!50!black] at (5.9, -1.55) {Memory: $O(nBk + Dk)$};

  \draw[thick, rounded corners] (8.5, -0.8) rectangle (11.4, 1.8);
  \node[font=\bfseries] at (9.95, 1.35) {Key result};
  \node[font=\scriptsize, align=center] at (9.95, 0.55) {\textcolor{green!50!black}{\textbf{60\% less}}\\activation memory};
  \node[font=\scriptsize, align=center] at (9.95, -0.25) {\textbf{No significant}\\accuracy loss};

\end{tikzpicture}
\caption{\textbf{LASER overview.} Standard TRM training stores full activations
$X_1, \dots, X_n$, with each $X_i \in \mathbb{R}^{B \times D}$, across recursive steps.
LASER instead stores compressed coefficients $Z_1, \dots, Z_n$, where
$Z_i = X_iQ \in \mathbb{R}^{B \times k}$, together with a shared low-rank basis
$Q \in \mathbb{R}^{D \times k}$, where $k \ll D$ is the retained rank.}
\label{fig:overview}
\end{figure}
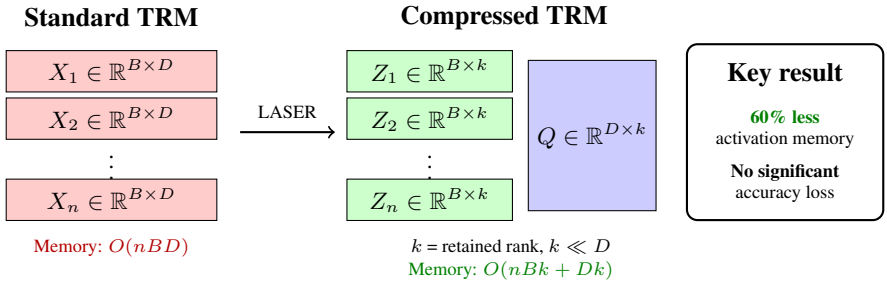

\section{Introduction}

Implicit reasoning, the ability to perform multi-step computation
through iterative latent processing rather than explicit
chain-of-thought generation, has emerged as a promising paradigm for both parameter-efficient and capable reasoning models. Tiny Recursive Models
(TRMs)~\citep{trm_less_is_more} exemplify this approach: a single
transformer block is unrolled over many recursive time-steps,
enabling deep computation with minimal parameters. While TRMs
achieve strong performance on structural reasoning tasks, their
training requires storing activations across all unrolled steps
for backpropagation through time, creating a memory bottleneck
that limits practical reasoning depth.

This bottleneck raises a natural question: what is the intrinsic
dimensionality of these stored reasoning trajectories? If
activations across 24 recursive steps span a high-dimensional,
nonlinearly curved manifold, then reducing memory requires
powerful but expensive methods. However, if the manifold is
low-rank and approximately linear, then the iterative reasoning
process is far more constrained than the architecture nominally
permits, and the memory problem becomes tractable.

We find the latter. TRM activations occupy an effectively linear,
low-dimensional subspace whose principal directions can be tracked
dynamically with cheap power iterations. This finding has both
interpretive and practical significance. Interpretively, it
suggests that weight-sharing concentrates iterative computation
along a small number of dominant eigendirections, raising
questions about the effective utilization of reasoning depth.
Practically, it motivates \textbf{LASER} (Low-Rank Activation SVD
for Efficient Recursion), a dynamic compression framework that
exploits this structure to achieve ${\sim}60\%$ activation memory
savings with negligible accuracy loss. 

This structure also suggests that recursive models are especially well-suited to activation compression. In standard backpropagation through time, activation memory grows directly with the number of recursive steps because a full hidden state must be retained at each step. LASER instead stores a compressed representation at each step together with a shared low-rank basis, so its advantage becomes larger as recursion depth increases. We evaluate this effect in the 24-step TRM setting used in prior work, both to remain comparable to the original architecture and to stay within our training budget.

Our contributions are:
\begin{itemize}
\item \textbf{Low-rank geometry:} TRM activations exhibit strong spectral decay and are well-approximated by a linear low-rank subspace.
\item \textbf{Adaptive tracking:} LASER tracks this evolving subspace with matrix-free power iteration and fidelity-triggered fallback.
\item \textbf{Practical compression:} LASER reduces activation memory by about 60\% with no statistically significant loss in performance.
\item \textbf{Quantization compatibility:} TRM activations remain stable under INT8 quantization after low-rank projection.
\end{itemize}
\section{Related Work}
\label{sec:relatedwork}
\subsection{Activation Checkpointing and Quantization}Standard memory-reduction techniques generally fall into two categories: \textit{Rematerialization} and \textit{Quantization}. Gradient Checkpointing (or Rematerialization) reduces memory by discarding intermediate activations and recomputing them during the backward pass. While effective, it imposes significant computational overhead when done aggressively, and we propose that LASER can be applied in parallel with non-aggressive checkpointing. Scalar quantization methods like {ActNN}~\citep{actnn} and {GIST}~\citep{GIST} compress activations by reducing precision (e.g., to 2-bit or 4-bit integers). While efficient, pure quantization treats activations as statistically independent scalars, ignoring the strong correlations between features. In this work, we demonstrate that quantization can complement subspace methods: LASER applies quantization to its low-rank \textit{residuals}, yielding higher compression rates than either method alone.
\subsection{Low-Rank Activation Compression}A more structurally aware approach to compression exploits the low-rank nature of neural activations. If the activation matrix $X \in \mathbb{R}^{N \times d}$ (where $N = B \times L$) resides in a subspace of rank $k \ll d$, it can be stored as two smaller factors $U \in \mathbb{R}^{N \times k}$ and $V \in \mathbb{R}^{d \times k}$.\subsubsection{Static and Randomized Projections}Recent work has attempted to apply this principle to training. {LANCE} ~\citep{lance_2025} employs a ``one-shot" Higher-Order SVD (HOSVD) to fix a compression basis at initialization. This eliminates decomposition overhead, but assumes the activation subspace at epoch 0 is the same as at epoch 100. Our experiments show this stationarity assumption fails for recursive models, where the manifold evolves as the model learns to reason. {GALE} ~\citep{gale_2025} uses randomized projections (sketching) to compress states. However, applying randomized sketching to activations at every layer introduces significant computational overhead.\subsubsection{Subspace Tracking}Our method draws inspiration from signal processing techniques for subspace tracking, such as the Projection Approximation Subspace Tracking (PAST) algorithm~\citep{PAST}. Unlike {LoRAct} ~\citep{loract_2025}, which uses sampling-based orthogonal decomposition, LASER applies a power-iteration scheme directly to the activation stream. Furthermore, we address the primary weakness of iterative tracking, error accumulation, by implementing an adaptive fallback: when the tracked subspace fails to explain the variance, we trigger a sparse, exact SVD update to realign the basis.

While the above methods focus on compression as an engineering goal, our work
additionally treats compressibility as a \emph{diagnostic}: the degree to which
activations admit low-rank approximation reveals structural properties of the
implicit reasoning process itself.
\section{Approach}
\subsection{The Geometry of Implicit Reasoning Trajectories}
\label{sec:manifold}

We perform principal component analysis on activations collected from trained
TRMs across all recursive time-steps. The spectral decay is rapid: at the MLP
intermediate site (dimension 3072), the top 128 components capture the vast
majority of activation variance (Figure~\ref{app:pca} in the Appendix). This implies that
out of 3072 available dimensions, the model's reasoning trajectory is
effectively confined to a subspace roughly 24$\times$ smaller.

To rule out the possibility that a nonlinear manifold merely \emph{appears}
linear under PCA, we compared linear projection against nonlinear autoencoder
baselines. The autoencoders provided no meaningful improvement in
reconstruction quality, confirming that the low-rank structure is genuinely
linear rather than a curved manifold that happens to be locally flat.

This linearity is not coincidental---it arises from architectural induction
biases. Smooth activations (GELU, SiLU) are locally linear over the
operating range; LayerNorm constrains activations to a bounded manifold; and
critically, recursive weight-sharing forces the same linear operator to be
applied repeatedly, concentrating the activation distribution along the
dominant eigendirections of the shared weight matrices. The result is that
implicit reasoning in TRMs, despite unrolling over many steps, proceeds along
a low-dimensional linear corridor.

\subsection{Activation Statistics and Quantization Feasibility}
\label{activationstats}
To evaluate whether TRM activations are suitable for low-precision compression, we analyzed their empirical distribution at both MLP and attention sites. Across 2.1M samples, activations were approximately zero-mean with near-Gaussian value histograms and tightly concentrated per-channel variances. Crucially, we observed no heavy-tailed behavior or frequent outliers.

These properties imply that TRM activations are highly compressible. Simulated INT8 quantization using symmetric scaling introduced negligible distortion (relative MSE $<0.03\%$), and applying quantization \emph{after} low-rank projection preserved means and standard deviations within $2\%$ across a range of ranks. 

Overall, TRM activations inhabit a well-behaved statistical regime—low-rank, nearly isotropic, and free of significant outliers—making them well-suited for LASER's low-rank–plus–quantization design. Full distributional analyses and quantization metrics are provided in Appendix~\ref{app:activationstats},~\ref{app:quanterror}.

\subsection{LASER Algorithm}
\label{sec:algorithm}
\begin{algorithm}[h]
\caption{LASER: Low-Rank Activation SVD for Efficient Recursion}
\begin{algorithmic}[1]
\REQUIRE Streams $\{X_t^{(s)}\}_{s \in \mathcal{S}}$, rank $k_0$, threshold $\varepsilon$, patience $p$, expansion $m$
\STATE \textbf{Init}: $\forall s$: $Q_0^{(s)} \leftarrow \text{top-}k_0\text{ SVD}(X_0^{(s)})$; $c^{(s)} \leftarrow 0$
\FOR{each training step $t$}
    \FOR{each site $s$}
        \STATE $Z \leftarrow X_t^{(s)} Q_{t-1}^{(s)}$ \hfill \textit{// Compress}
        \STATE $F \leftarrow \|Z\|_F / \|X_t^{(s)}\|_F$ \hfill \textit{// Fidelity}
        
        \IF{$F \geq \varepsilon$}
            \STATE $Q_t^{(s)} \leftarrow \text{Orth}(X_t^{(s)T} Z)$ \hfill \textit{// Power iteration}
            \STATE $c^{(s)} \leftarrow 0$
        \ELSE
            \STATE $c^{(s)} \leftarrow c^{(s)} + 1$
            \IF{$c^{(s)} \geq p$}
                \STATE $Q_t^{(s)} \leftarrow \text{top-}k_0\text{ SVD}(X_t^{(s)})$ \hfill \textit{// Reset}
                \STATE $c^{(s)} \leftarrow 0$
            \ELSE
                \STATE Sample $m$ rows from $X_t^{(s)}$ as $R$ \hfill \textit{// Expand}
                \STATE $R_\perp \leftarrow R - R Q_{t-1}^{(s)} Q_{t-1}^{(s)T}$
                \STATE $Q_t^{(s)} \leftarrow \text{Orth}([Q_{t-1}^{(s)} | R_\perp^T])$
            \ENDIF
        \ENDIF
    \ENDFOR
    \STATE \textbf{Backward}: $\forall s$: use $\hat{X}^{(s)} = Z^{(s)} Q_t^{(s)T}$
\ENDFOR
\end{algorithmic}
\label{alg:laser}
\end{algorithm}
LASER approximates the activation matrix $X \in \mathbb{R}^{B \times D}$ using a low-rank basis $Q \in \mathbb{R}^{D \times k}$, such that $X \approx X Q Q^T$. We store the compressed representation $Z = XQ \in \mathbb{R}^{B \times k}$ and the basis $Q$, reducing memory complexity from $O(BD)$ to $O(Bk + Dk)$.

Instead of computing the expensive covariance matrix $X^T X$ for every batch (an $O(D^2)$ operation), LASER updates the basis $Q$ using a matrix-free power iteration step. The new subspace direction is estimated via $M = X^T (X Q)$. This is computed in $O(BDk)$, significantly faster than full SVD. $M$ is then orthonormalized to produce the updated basis $Q_{new}$.

A major failure mode of iterative tracking is ``subspace drift,'' where the basis $Q$ slowly diverges from the true principal components during rapid distribution shifts or poor conditioning. LASER mitigates this via a fidelity-based response mechanism.

We define a fidelity metric $F_t = ||Z_t||_F / ||X_t||_F$. If $F_t < \varepsilon$ (a pre-defined threshold), LASER first attempts a subspace expansion. $m$ activation vectors are sampled, orthogonalized against the current basis, and appended to $Q$, immediately broadening the subspace. If fidelity remains low for $p$ consecutive batches, LASER triggers a ``Hard Reset.'' We compute the exact SVD on the current micro-batch to perfectly realign $Q$ with the current activation variance, but do not backpropagate over that batch to avoid a memory spike. This hybrid approach allows for the speed of power iterations with the stability of exact SVD.

LASER maintains separate bases $Q^{(s)}$ for each compression site $s$ (e.g., 
MLP intermediate, MLP output, attention output). Each site tracks its own 
fidelity $F_t^{(s)}$ and adapts rank independently. This allows aggressive 
compression where possible while 
preserving capacity where needed (in particular, later-stage activations often require higher $k$).

LASER's key advantage over prior work is \textit{matrix-free adaptive tracking with graduated fallback}. Unlike methods outlined in Section~\ref{sec:relatedwork}, LASER tracks the evolving subspace via $O(BDk)$ power iteration, only invoking exact SVD when fidelity drops persistently. This yields the efficiency of static methods with the robustness of dynamic ones.

\subsection{Memory Scaling with Recursion Depth}
\label{sec:scaling}

LASER is especially well-suited to recursive architectures because its memory advantage grows with recursion depth. For recursion depth $n$, batch size $B$, activation width $D$, and retained rank $k \ll D$, standard backpropagation stores activations with memory
\[
M_{\text{full}}(n)=O(nBD),
\]
whereas LASER stores compressed per-step coefficients together with a shared basis:
\[
M_{\text{LASER}}(n)=O(nBk + Dk).
\]
Thus LASER replaces the dominant per-step memory term in width $D$ with one in width $k$. As recursion depth increases, the one-time basis cost becomes negligible and the memory savings grow approximately linearly with $n$, making deeply recursive models a particularly favorable regime for low-rank activation compression.

\subsection{Theoretical Guarantees}
\label{sec:theory}
\subsubsection{Fidelity Metric}
The fidelity $F_t = \|Z\|_F / \|X\|_F$ exactly equals the cosine similarity between $X$ and its reconstruction $\hat{X} = ZQ^T$ when $Q$ is orthonormal (Proposition 1, Appendix~\ref{app:fidelitymetricjustification}). This enables zero-cost quality monitoring without reconstructing $\hat{X}$.

\subsubsection{Gradient Fidelity}
Reconstruction accuracy alone is insufficient—we require that gradients computed at $\hat{a} = QQ^Ta$ approximate those at $a$. Under standard smoothness assumptions (Lipschitz Jacobian with constant $L_J$), we prove:

\noindent\textbf{Theorem 1.} \textit{The gradient error satisfies $\|\tilde{g}_a - g_a\| \leq L_J \|\lambda\| \varepsilon$, where $\varepsilon = \|\hat{a} - a\|$ and $\lambda$ is the upstream gradient.}

\noindent\textbf{Corollary 1.} \textit{Gradient cosine similarity satisfies $\cos(g_a, \tilde{g}_a) \geq 1 - 2L_J\|\lambda\|\varepsilon / \|g_a\|$.}

The smoothness assumptions hold because TRM activations traverse a well-conditioned manifold: GELU/SiLU activations are $C^\infty$, LayerNorm bounds activation norms, and recursive weight-sharing concentrates activations along stable eigendirections. Together, these architectural choices ensure small $L_J$, meaning gradient fidelity degrades gracefully with reconstruction error. Crucially, LASER uses the \textit{full} Jacobian—only the evaluation point shifts. This avoids rank collapse that would occur if gradients flowed through a bottleneck layer. Proofs are in Appendix~\ref{app:gradient_fidelity}.
\section{Implementation}
We implemented LASER in PyTorch using forward/backward hooks, allowing it to be slotted into other models as well, theoretically.

\subsection{Experimental Setup}

We evaluate LASER on 11$\times$11 maze pathfinding, where a TRM must predict the solution path connecting start to goal. The dataset contains 10,000 procedurally generated mazes (9,000 train / 1,000 validation). Our TRM uses hidden dimension 512, 8 attention heads, SwiGLU MLP with 4$\times$ expansion, and 24 recursive L-cycles with gradient. We use 24 recursions to match the standard TRM setting from prior work; this depth is also sufficient to expose the activation-memory bottleneck while remaining feasible within our compute budget. LASER compresses all activations saved for backpropagation. Adaptive rank growth allows smaller tensors to recover full rank when needed, while large tensors (\texttt{mlp\_3072}, \texttt{attn\_1536}) remain highly compressed (57--92\%), providing the bulk of memory savings. We believe smaller tensors can also be compressed with tuning of LASER's hyperparameters. We compare baseline training against LASER with initial ranks $k \in \{128, 256\}$, reporting token accuracy and maze solve rate over 5 seeds. Full hyperparameters are provided in the Appendix, as well as seeds, for reproducibility.

\section{Results}
\label{sec:results}
\begin{figure}[h!]
\centering
\includegraphics[width=\linewidth]{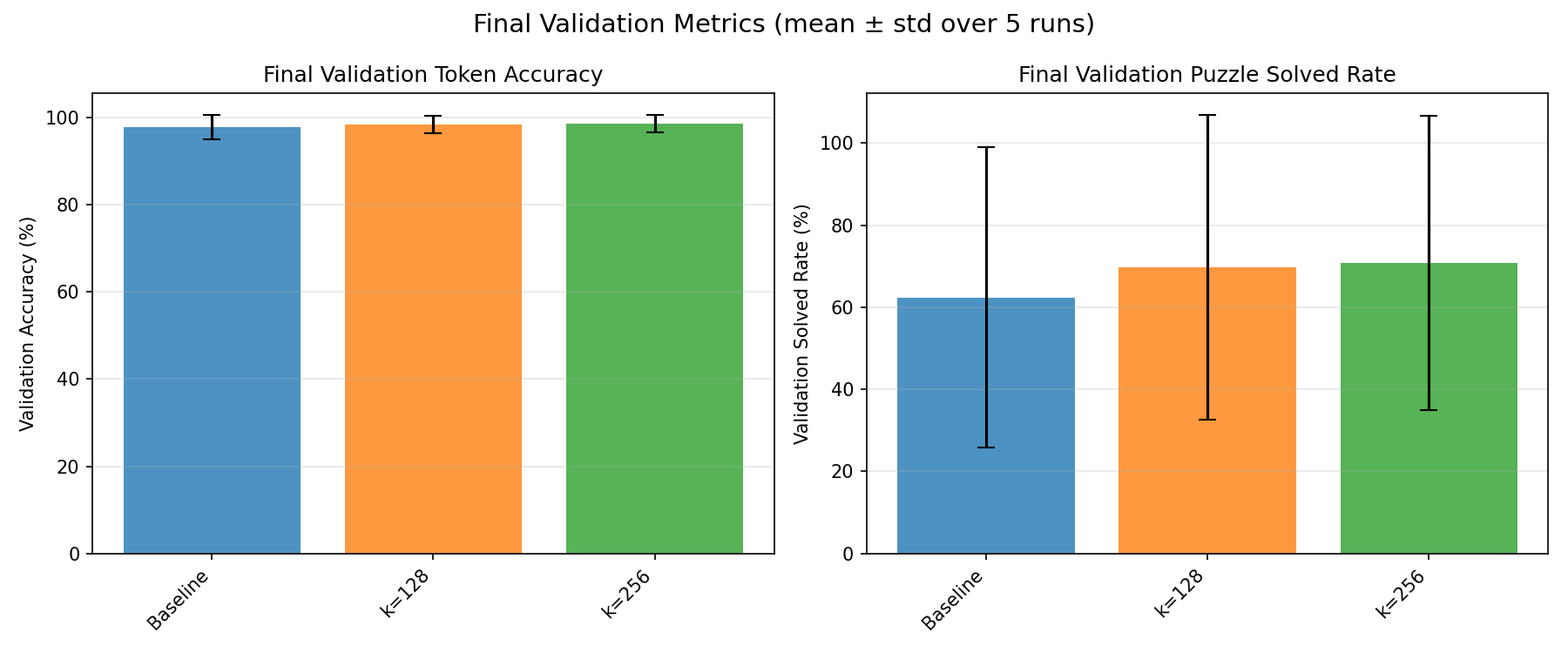}
\caption{\textbf{Validation performance under LASER compression.}
Across 24 recursive steps, LASER closely matches baseline validation behavior while reducing activation memory by approximately 60\%.}
\label{fig:val_metrics}
\end{figure}

Figure~\ref{fig:val_metrics} shows validation performance during training.
LASER closely tracks the baseline while using substantially less activation memory.

Table~\ref{tab:results} summarizes the final performance across five random
seeds. LASER achieves \textbf{60\% activation memory reduction}
(2.85 GB $\rightarrow$ 1.14 GB) with no statistically significant performance
degradation.

\begin{table}[h]
\centering
\caption{Performance (mean $\pm$ std, 5 seeds).}
\label{tab:results}
\begin{tabular}{@{}lccc@{}}
\toprule
\textbf{Method} & \textbf{Val Acc (\%)}  & \textbf{Val Solved (\%)} & \textbf{Act. Mem.} \\
\midrule
Baseline & 97.82$\pm$2.78 & 62.34$\pm$36.55 & 2.85 GB \\
LASER $k$=256 & 98.54$\pm$2.00 & 70.74$\pm$35.88 & 1.23 GB \\
LASER $k$=128 & 98.46$\pm$2.01 & 69.71$\pm$37.09 & 1.14 GB \\
\bottomrule
\end{tabular}
\end{table}

The primary compression target is \texttt{mlp\_3072}, achieving 92.5\% compression at $k$=128 and accounting for 65\% of total memory savings.

\begin{table}
\centering
\caption{Per-site compression (final ranks at $k$=128)}
\begin{tabular}{@{}lccc@{}}
\toprule
\textbf{Site} & \textbf{Dim} & \textbf{Final Rank} & \textbf{Savings} \\
\midrule
mlp\_3072 & 3072 & 128 & 92.5\% \\
attn\_1536 & 1536 & 470 & 57.3\% \\
mlp\_1536 & 1536 & 510 & 53.6\% \\
attn\_512 & 512 & 510 & 0\% (overhead) \\
mlp\_512 & 512 & 500 & 0\% (overhead) \\
attn\_64 & 64 & 64 & 0\% \\
\bottomrule
\end{tabular}
\label{sitespecific}
\end{table}

\section{Discussion}
\label{sec:discussion}

\paragraph{Where does redundancy live in implicit reasoning?}
The most striking finding is not the aggregate compression ratio
but its asymmetry across computational sites. The 3072-dimensional
MLP intermediate activations compress to rank 128---a 92.5\%
reduction---while the 512-dimensional hidden state and smaller
attention sites grow back toward full rank under LASER's adaptive
mechanism. This suggests that redundancy in implicit reasoning is
highly localized: the expanded MLP bottleneck is massively
over-parameterized relative to what each recursive step actually
computes, while the core hidden state genuinely utilizes its full
capacity. Weight-sharing appears to concentrate MLP computation
along a few dominant eigendirections, but this concentration is
site-specific rather than uniform.

\paragraph{Concentrated computation and reasoning efficiency.}
The rapid spectral decay at high-dimensional sites raises a
natural architectural question: if MLP intermediate activations
at each recursive step operate in a ${\sim}128$-dimensional
corridor within a 3072-dimensional space, it may be possible to
amplify computation along these directions and achieve comparable
reasoning with fewer but more effective recursive steps. The high
variance in maze solve rates (Table~\ref{tab:results}) hints that
current TRM training may not reliably find these productive
directions, and that methods which explicitly encourage signal
concentration, whether through architectural changes or training
objectives, could improve both efficiency and stability.

\paragraph{Smooth subspace evolution and reasoning complexity.}
LASER's success depends not only on the subspace being low-rank
but on it evolving smoothly enough for power iteration to track
during training. The fact that a single power iteration step per
batch suffices, with hard resets rarely triggered, implies that
the reasoning manifold at compressible sites changes gradually.
This is consistent with TRMs learning smooth, stereotyped
reasoning trajectories rather than chaotic or rapidly shifting
computation patterns. Whether this smoothness is a feature
(stable reasoning) or a limitation (inability to perform diverse
reasoning strategies) is an open question.

The $11\times11$ maze pathfinding task provides a highly structured environment, which likely represents a best-case scenario for smooth subspace evolution.

If LASER were applied to more complex, open-ended tasks where implicit reasoning requires diverse and rapidly shifting strategies, the activation subspace may exhibit much higher variance. In volatile tasks, rapid distribution shifts would cause ``subspace drift" where our iteratively tracked basis $Q$ diverges from the true principal components. If the activation space is so chaotic that this expansion fails to restore fidelity for $p$ consecutive batches, LASER abandons the power iteration and computes an exact SVD on the current micro-batch to realign the basis. Investigating how subspace dimensionality and drift correlate with task complexity remains a critical next step.

\textbf{Task dependence and the maze pathfinding regime.} The $11\times11$ maze pathfinding
task exhibits several properties that are particularly favorable for low-rank activation
compression: a small, fixed vocabulary (5 tokens), deterministic targets (unique shortest
paths), a stationary input distribution, and a uniform reasoning strategy (path search
applied identically to every input). Together, these properties mean the model converges to
a single stable computational routine, and the activation subspace reflects this, settling
into a low-dimensional corridor because the same computation is performed on every input
with only spatial configuration varying.

On tasks requiring diverse reasoning strategies, such as mathematical problem-solving,
natural language inference, or multi-step planning with branching, we would expect the
activation subspace to be higher-dimensional (encoding multiple computational pathways),
less stable across inputs (different problems activating different directions), and potentially
non-stationary across training as the model acquires qualitatively new capabilities. LASER's
adaptive mechanisms, namely rank expansion and fidelity-triggered resets, are designed to
accommodate such variation, but whether power iteration can track a rapidly shifting
subspace without frequent hard resets remains an open empirical question. We note that the
overhead of hard resets is bounded (one exact SVD per micro-batch per site), so even in a
volatile regime LASER degrades gracefully to periodic exact decomposition rather than
failing silently.

\paragraph{Implications for other latent reasoning architectures.}
Our analysis is specific to TRMs, but the mechanism we believe
drives compressibility (repeated application of shared weights
concentrating activations along dominant eigendirections) should
apply to any looped or weight-tied architecture. We conjecture
that similar site-specific low-rank structure exists in other
implicit reasoning systems (e.g., Universal Transformers), and
that the effective dimensionality of activations at different
computational sites could serve as a probe for where redundancy
and capacity are allocated in latent reasoning.

\paragraph{Applicability to LoopLM and larger recursive models.}
LoopLM \citep{looplm2025} is a particularly relevant test case, as it
applies the same weight-tied recursive principle as TRMs but at
significantly larger scale (1.4B--2.6B parameters) and on natural
language rather than a structured 5-token vocabulary. The core
mechanism we identify, recursive weight-sharing concentrating
activations along dominant eigendirections, does not depend on model
scale; it depends on the spectral properties of the shared weights.
MLP intermediate layers are widely observed to be over-parameterized
in standard (non-recursive) transformers
\citep{gromov2024unreasonable}, and weight-sharing should produce the
same concentrating effect regardless of hidden dimension. The
compression-relevant quantity is the ratio of effective rank to
ambient dimension, which we expect to remain favorable.

However, two factors could reduce compressibility in this setting.
First, models trained on natural language exhibit heavier-tailed
activation distributions with more frequent outliers than we observe
in the maze vocabulary (Section~\ref{activationstats}), which could
degrade both low-rank approximation quality and post-projection
quantization. Second, richer tasks may require higher effective rank
at each computational site, narrowing the gap between $k$ and $D$
and reducing LASER's memory advantage. Validating whether the
site-specific compression asymmetry we observe
(Table~\ref{sitespecific}), with MLP intermediates highly compressible
and hidden states near full rank, persists in LoopLM and similar
architectures remains to be tested empirically.

\subsection{Comparisons with Alternative Methods in Literature}

While computational and time constraints prevent us from benchmarking LASER directly against existing activation compression methods in the TRM setting, we identify principled structural differences that distinguish the recursive regime from the settings these methods were designed for. These differences partially motivated our development of LASER.

\textbf{ActNN} \citep{actnn} achieves up to 12$\times$ activation memory reduction through mixed-precision scalar quantization, treating each activation element independently. This approach is \textbf{orthogonal} to LASER's subspace compression: ActNN removes bit-width redundancy while LASER removes structural redundancy arising from weight-sharing. Our quantization analysis (Section \ref{activationstats}, Appendix \ref{app:activationstats}, \ref{app:quanterror}) demonstrates that these two compression axes compose well: INT8 quantization applied after low-rank projection introduces negligible additional distortion ($<$0.03\% relative MSE). We therefore view ActNN-style quantization as complementary to LASER rather than competitive. 

\textbf{LANCE} \citep{lance_2025} fixes a low-rank basis at initialization via one-shot HOSVD and reuses it throughout training, which is an efficient strategy for fine-tuning pre-trained models where the activation subspace is already near equilibrium. However, for training that proceeds from random initialization, such as our TRM training setup here, the activation subspace at epoch 0 bears little resemblance to the converged reasoning subspace. Our PCA analysis (Figure \ref{app:pca}) reveals that even when the SVD basis is \textit{recomputed exactly at every training step}, an oracle upper bound on any fixed-rank projection, lower-rank reconstructions exhibit a pronounced mid-training fidelity dip. Since the basis is optimal at each step, this dip cannot reflect directional misalignment; rather, it indicates that the effective dimensionality of the activation manifold expands as the model transitions between learning regimes. A static basis like LANCE's would face this same dimensionality expansion, which LASER accounts for, \textit{and additionally} suffer from directional staleness: using step-0 principal components that are unlikely to span the relevant subspace at later training stages. This second source of error is invisible in the oracle curves, meaning Figure \ref{app:pca} represents a generous upper bound on LANCE's achievable fidelity. LASER addresses both failure modes: adaptive rank growth accommodates dimensionality expansion, while power iteration tracks directional drift.

\textbf{LoRAct} \citep{loract_2025} is the most structurally similar method to LASER, performing online low-rank decomposition via sampling-based orthogonal factorization. However, LoRAct was designed for parameter-efficient fine-tuning of large pre-trained models, where each layer has distinct weights producing genuinely different activation distributions that warrant independent decomposition. In a 24-step TRM with shared weights, the same operator is applied at every recursive step, concentrating activations along the same dominant eigendirections. LoRAct would perform 24 independent decompositions per site that each converge to approximately the same subspace, while LASER amortizes a single shared basis $Q$ across all steps, a distinction whose computational advantage grows linearly with recursion depth. Additionally, because LoRAct recomputes the decomposition from scratch rather than tracking incrementally, it avoids subspace staleness entirely, at the cost of having to recompute fully at every step.

\textbf{GALE} \citep{gale_2025} applies randomized sketching to compress activations, using data-independent random projections that preserve geometric structure. This data-independence means GALE cannot exploit the pronounced low-rank structure that weight-sharing induces: it treats the 128-dimensional effective subspace within a 3072-dimensional MLP identically to a full-rank distribution. However, this same property makes GALE immune to subspace drift by construction. On tasks where the activation manifold shifts rapidly and LASER's iterative tracking would require frequent hard resets, GALE's fixed random projections would maintain consistent compression quality. The per-step overhead of sketching is higher than LASER's power iteration and scales linearly with recursion depth, but this cost may be justified in volatile regimes. Investigating the tradeoff between structure-exploiting methods like LASER and structure-agnostic methods like GALE across tasks of varying complexity is a promising direction for future work.


\section{Conclusion}

We introduced LASER, a dynamic compression framework for training recursive
models, and used it to reveal that TRM reasoning trajectories occupy a
low-dimensional linear subspace that evolves smoothly during training. LASER
achieves over 60\% activation memory savings with accuracy statistically
indistinguishable from baseline, supported by theoretical gradient fidelity
guarantees.

Our analysis raises open questions about the effective reasoning depth of
recursive architectures and the relationship between subspace dimensionality
and reasoning complexity. Limitations include evaluation on a single task
(maze pathfinding) and the need for manual tuning of adaptive rank growth.
Future work includes extending this analysis to other looped architectures,
investigating whether subspace dimensionality correlates with task difficulty,
and combining LASER with gradient checkpointing for further memory reduction.

\subsubsection*{Author Contributions}
{Ege \c{C}akar}: LASER theory development, LASER implementation and updates, activation space linearity and autoencoder ablations, LASER ablations, additional ablations, literature comparison, transfer across architectures analysis, experimental debugging, and final manuscript formatting. 

{Lia Zheng}: Quantization analysis (including outlier characterization), figure editing, theoretical scaling law edits, additional ablations, and experimental debugging. 

{Ketan Raghu}: Activation quantization ablations, exploratory ablations on combined LASER and quantization, additional ablations, and experimental debugging.

\newpage
\bibliographystyle{iclr2026_conference}
\bibliography{iclr2026_conference}
\newpage
\appendix

\section{Appendix}

\subsection{Fidelity Metric Justification}
\label{app:fidelitymetricjustification}
We formally demonstrate that $F_t$ is exactly equivalent to the cosine similarity between the original activations and their low-rank reconstruction, provided the basis is orthonormal.

\noindent \textbf{Proposition 1.} \textit{Let $X \in \mathbb{R}^{N \times D}$ be the activation matrix and $Q \in \mathbb{R}^{D \times k}$ be a semi-orthogonal basis such that $Q^T Q = I_k$. Let $Z = XQ$ be the compressed state and $\hat{X} = ZQ^T$ be the reconstruction. Then, the fidelity ratio equals the cosine similarity: $\frac{\|Z\|_F}{\|X\|_F} = \text{sim}(X, \hat{X})$.}

\noindent \textit{Proof.} Recall the definition of cosine similarity for matrices: $\text{sim}(X, \hat{X}) = \frac{\langle X, \hat{X} \rangle_F}{\|X\|_F \|\hat{X}\|_F}$. First, we simplify the inner product (numerator) using the cyclic property of the trace:
\begin{equation}
\begin{aligned}
\langle X, \hat{X} \rangle_F &= \text{Tr}(X^T (X Q Q^T)) = \text{Tr}(Q^T X^T X Q) \\
&= \text{Tr}((XQ)^T (XQ)) = \|Z\|_F^2
\end{aligned}
\end{equation}
Next, we determine the norm of the reconstruction (denominator). Since $Q$ is orthonormal, the projection preserves the norm of the coefficients $Z$:
\begin{equation}
\begin{aligned}
\|\hat{X}\|_F^2 &= \text{Tr}((ZQ^T)^T (ZQ^T)) = \text{Tr}(Q Z^T Z Q^T) \\
&= \text{Tr}(Z^T Z (Q^T Q)) = \|Z\|_F^2 \implies \|\hat{X}\|_F = \|Z\|_F
\end{aligned}
\end{equation}
Substituting these results back into the similarity definition:
\begin{equation}
\text{sim}(X, \hat{X}) = \frac{\|Z\|_F^2}{\|X\|_F \|Z\|_F} = \frac{\|Z\|_F}{\|X\|_F} \equiv F_t \quad \blacksquare
\end{equation}
This equivalence ensures that monitoring $F_t$ provides a mathematically rigorous measure of angular alignment without requiring the expensive $O(ND)$ reconstruction of $\hat{X}$.

\subsection{Gradient Fidelity Guarantees}
\label{app:gradient_fidelity}

Reconstruction accuracy is necessary but not sufficient: we need gradients computed at reconstructed activations to approximate true gradients.

\subsubsection{Setup}
Let $a \in \mathbb{R}^D$ be the activation at a compression site, and let $f(a, \theta): \mathbb{R}^D \to \mathbb{R}^m$ represent the downstream computation. Let $\hat{a} = QQ^T a$ be the reconstructed activation with error $\varepsilon = \|\hat{a} - a\|$. The true gradient is $g_a = J_a f(a, \theta)^T \lambda$ where $\lambda = \nabla \mathcal{L}$ is the upstream gradient; LASER computes $\tilde{g}_a = J_a f(\hat{a}, \theta)^T \lambda$.

\subsubsection{Assumptions}
We require smoothness conditions satisfied by networks with GELU/SiLU activations and LayerNorm:
\begin{enumerate}
    \item \textbf{Lipschitz Jacobian}: $\|J_a f(a') - J_a f(a)\| \leq L_J \|a' - a\|$
    \item \textbf{Bounded Jacobian}: $\|J_a f(a)\| \leq M_a$
\end{enumerate}

\subsubsection{Main Result}

\noindent \textbf{Theorem 1 (Gradient Error Bound).} \textit{Under the Lipschitz Jacobian assumption:}
\begin{equation}
    \|\tilde{g}_a - g_a\| \leq L_J \|\lambda\| \varepsilon
\end{equation}

\noindent \textit{Proof.}
$\|\tilde{g}_a - g_a\| = \|(J_a f(\hat{a}) - J_a f(a))^T \lambda\| \leq \|J_a f(\hat{a}) - J_a f(a)\| \|\lambda\| \leq L_J \varepsilon \|\lambda\|$ \hfill $\blacksquare$

\noindent \textbf{Corollary 1 (Cosine Alignment).} \textit{The gradient cosine similarity satisfies:}
\begin{equation}
    \cos(g_a, \tilde{g}_a) \geq 1 - \frac{2 L_J \|\lambda\| \varepsilon}{\|g_a\|}
\end{equation}

\subsubsection{Interpretation}
This result is crucial: LASER uses the \textit{full} Jacobian $J_a f$, only the evaluation point shifts from $a$ to $\hat{a}$. This avoids the rank collapse that would occur if gradients flowed through a $k$-dimensional bottleneck. Combined with the Eckart-Young-Mirsky theorem guaranteeing SVD optimality, this explains why LASER outperforms autoencoders despite their greater representational capacity.

\subsection{Activation Statistics and Quantization Details}
\label{app:activationstats}

We characterized activation distributions at the MLP and attention compression sites using 2.1M activation samples collected over 425 batches. Both sites exhibited approximately Gaussian value histograms with near-zero means and per-dimension variances of $0.91$ (MLP) and $1.46$ (attention). Per-channel standard deviations were tightly concentrated in the ranges $[0.7, 1.1]$ for MLP and $[0.9, 1.4]$ for attention.

Outlier rates closely matched Gaussian predictions. For MLP activations we observed
\[
\mathbb{P}(|x-\mu| > 2\sigma) = 4.1\times 10^{-2}, \quad
\mathbb{P}(|x-\mu| > 3\sigma) = 3.0\times 10^{-3},
\]
with only $1.4\times10^{-4}$ and $6.7\times10^{-6}$ samples exceeding $4\sigma$ and $5\sigma$, respectively. Attention activations displayed similar behavior.

We simulated INT8 quantization using (i) max-based scaling and (ii) $4\sigma$-based scaling. For both MLP and attention sites, max-based scaling yielded a relative MSE of approximately $1.4\times10^{-4}$, while $4\sigma$ scaling produced at most $2.9\times10^{-4}$. Thus, quantization noise accounts for less than $0.03\%$ of activation energy.

Finally, applying INT8 quantization after PCA projection at ranks $k \in \{64, 128, 256\}$ produced reconstructed activation statistics nearly identical to the raw distributions: means and standard deviations changed by less than $2\%$, and overlaid histograms were visually indistinguishable.

\subsection{Quantization Error}
\label{app:quanterror}
\begin{figure}[H]
    \centering
    \includegraphics[width=0.5\linewidth]{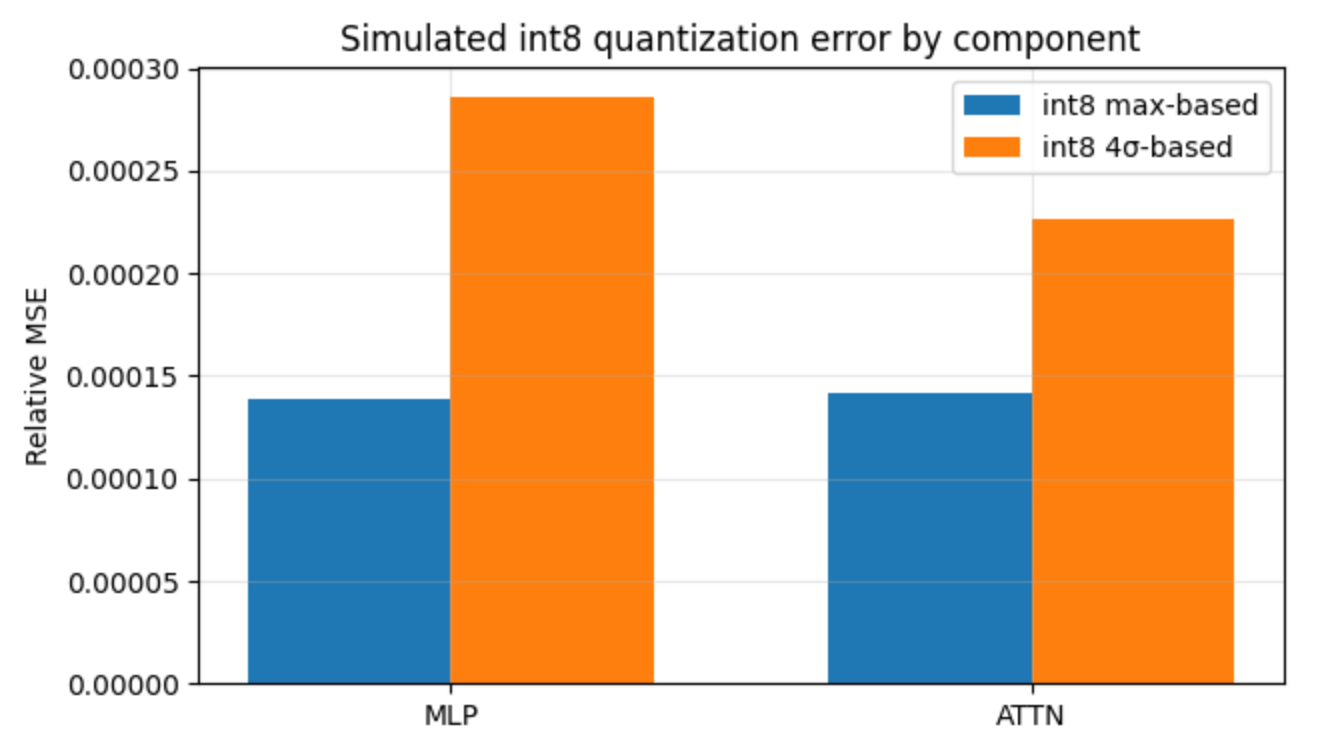}
    \caption{Simulated error by component}
    \label{fig:quant_mse}
\end{figure}

We evaluate post-hoc INT8 activation quantization at the MLP and attention sites using the two scaling schemes from Section~\ref{activationstats}. The resulting relative MSE is extremely small: $\approx 1.4\times 10^{-4}$ with max-based scaling and at most $2.9\times 10^{-4}$ with $4\sigma$ scaling for both sites (Figure~\ref{fig:quant_mse}). Applying quantization after PCA projection with ranks $k\in\{64,128,256\}$ yields nearly identical marginal statistics to the raw activations (means and standard deviations change by $<2\%$), and overlaid histograms are visually indistinguishable. Thus, in the TRM regime we study, low-rank projection plus INT8 quantization introduces only a tiny perturbation, supporting the use of LASER as a low-rank+quantization scheme for activation memory reduction.

\begin{table}[h]
\centering
\caption{Model configuration}
\begin{tabular}{@{}ll@{}}
\toprule
Hidden size & 512 \\
Attention heads & 8 (head dim 64) \\
MLP expansion & 4.0$\times$ (SwiGLU, inter=1536) \\
H-cycles / L-cycles & 1 / 24 \\
Position encoding & RoPE ($\theta$=10000) \\
Parameters & $\sim$3.4M \\
\bottomrule
\end{tabular}
\end{table}

\begin{table}[h]
\centering
\caption{Training configuration}
\begin{tabular}{@{}ll@{}}
\toprule
Optimizer & AdamW ($\beta$=(0.9, 0.95), wd=$10^{-2}$) \\
Learning rate & $10^{-4}$, cosine decay \\
Batch size & 64 \\
Epochs & 16 \\
Gradient clip & 1.0 \\
Precision & bfloat16 (AMP) \\
Seeds & 100--104 (5 runs) \\
\bottomrule
\end{tabular}
\end{table}

\begin{table}[h]
\centering
\caption{LASER configuration}
\begin{tabular}{@{}ll@{}}
\toprule
Initial rank ($k$) & 128 or 256 \\
Fidelity threshold ($\varepsilon$) & 0.95 \\
Growth size ($m$) & 4 \\
Max rank & 512 \\
Power iteration steps & 1 \\
Reset patience ($p$) & 2 \\
\bottomrule
\end{tabular}
\end{table}

\begin{table}[h]
\centering
\caption{Dataset: 11$\times$11 maze pathfinding}
\begin{tabular}{@{}ll@{}}
\toprule
Samples & 10,000 (9k train / 1k val) \\
Grid size & 11$\times$11 (seq\_len=121) \\
Vocabulary & 5 tokens (wall, passage, start, goal, path) \\
Task & Predict shortest path from start to goal \\
\bottomrule
\end{tabular}
\end{table}

\begin{figure}[H]
    \centering
    \includegraphics[width=0.8\linewidth]{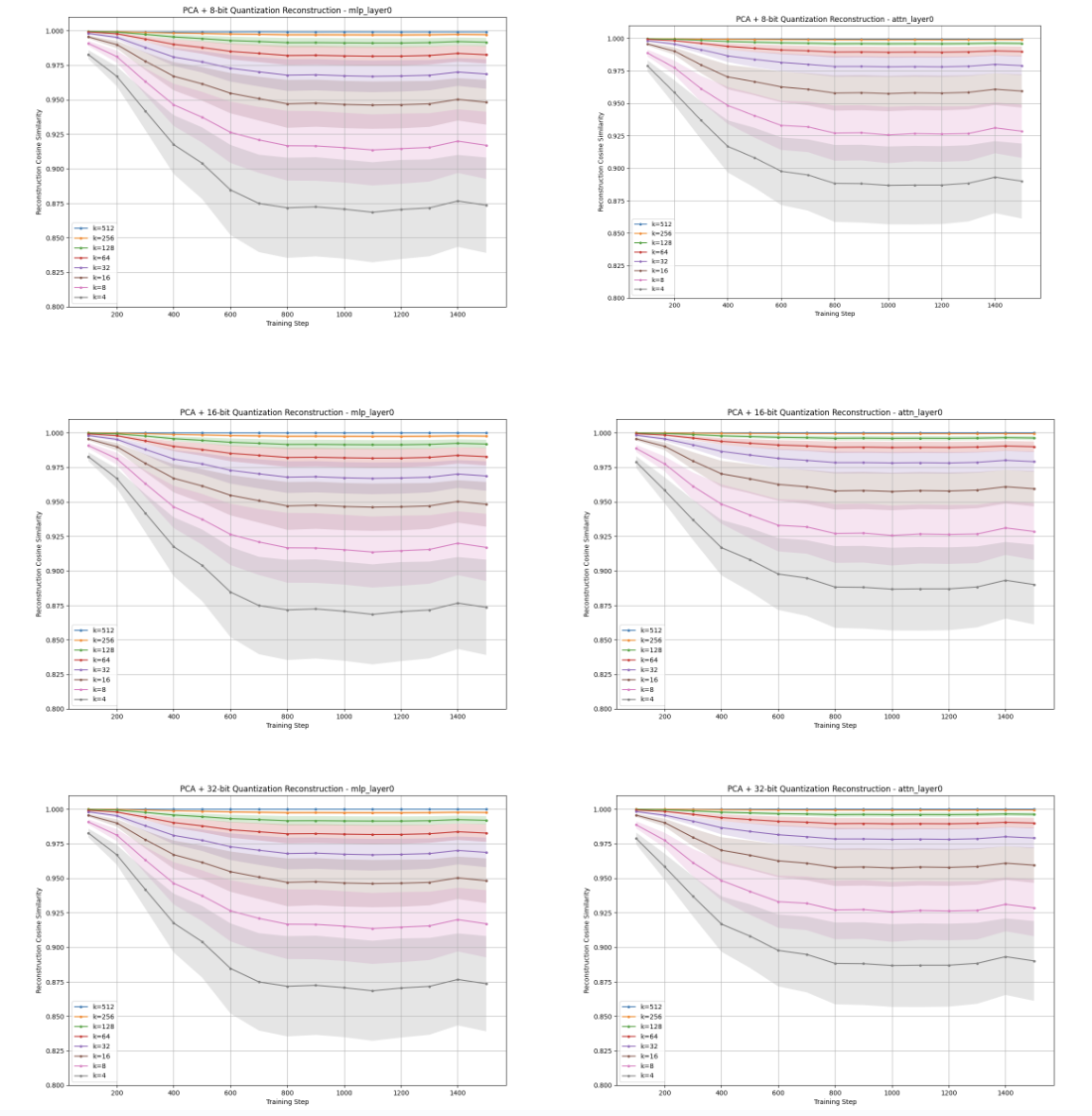}
    \caption{PCA reconstruction cosine similarity across training
    steps at the MLP (left) and attention (right) sites under 8-bit
    (top), 16-bit (middle), and 32-bit (bottom) quantization, with
    the SVD basis recomputed exactly each batch. Rank $k{=}128$
    maintains $>$0.975 fidelity for the MLP activations throughout
    training, confirming rapid spectral decay. Quantization precision
    has minimal effect on reconstruction quality. The mid-training
    fidelity dip reflects the evolving activation subspace as the
    model learns, motivating LASER's adaptive tracking over static
    bases.}
    \label{app:pca}
\end{figure}

\end{document}

%% file: math_commands.tex
\usepackage{amsmath,amsfonts,bm}

\def\eqref#1{equation~\ref{#1}}

\def\1{\bm{1}}

\DeclareMathAlphabet{\mathsfit}{\encodingdefault}{\sfdefault}{m}{sl}
\SetMathAlphabet{\mathsfit}{bold}{\encodingdefault}{\sfdefault}{bx}{n}

